\documentclass[conference]{IEEEtran}
\IEEEoverridecommandlockouts
\renewcommand{\baselinestretch}{1}
\usepackage{cite}
\usepackage{amsmath,amssymb,amsfonts}
\usepackage{graphicx}
\usepackage{textcomp}
\usepackage{algpseudocode}
\usepackage{comment}
\usepackage{algorithm}
\usepackage{xcolor}
\usepackage{comment}
\usepackage{multirow}
\usepackage{url}
\usepackage{hyperref}
\usepackage{eso-pic}
\usepackage{makecell}
\usepackage{subcaption}
\usepackage{booktabs}

\def\BibTeX{{\rm B\kern-.05em{\sc i\kern-.025em b}\kern-.08em
    T\kern-.1667em\lower.7ex\hbox{E}\kern-.125emX}}

\title{Optimizing Stroke Risk Prediction: A Machine Learning Pipeline Combining ROS-Balanced Ensembles and XAI}
\author{
\IEEEauthorblockN{
A S M Ahsanul Sarkar Akib\IEEEauthorrefmark{1},
Raduana Khawla\IEEEauthorrefmark{2},
Abdul Hasib\IEEEauthorrefmark{3}
}

\IEEEauthorblockA{
\IEEEauthorrefmark{1}Department of Robotics, Robo Tech Valley, Dhaka, Bangladesh\\
\IEEEauthorrefmark{2}Department of Computer Science and Engineering, Jagannath University, Dhaka, Bangladesh\\
\IEEEauthorrefmark{3}Department of Internet of Things \& Robotics Engineering, University of Frontier Technology, Bangladesh
}

\IEEEauthorblockA{
Emails:
\IEEEauthorrefmark{1}ahsanulakib@gmail.com,
\IEEEauthorrefmark{2}khawla.cse.jnu@gmail.com,
\IEEEauthorrefmark{3}sm.abdulhasib.bd@gmail.com
}
}
\begin{document}
\maketitle
\begin{abstract}
Stroke is a major cause of death and permanent impairment, making it a major worldwide health concern. For prompt intervention and successful preventative tactics, early risk assessment is essential.
To address this challenge, we used ensemble modeling and explainable AI (XAI) techniques to create an interpretable machine learning framework for stroke risk prediction. A thorough evaluation of 10 different machine learning models using 5-fold cross-validation across several datasets was part of our all-inclusive strategy, which also included feature engineering and data pretreatment (using Random Over-Sampling (ROS) to solve class imbalance).
Our optimized ensemble model (Random Forest + ExtraTrees + XGBoost) performed exceptionally well, obtaining a strong 99.09\% accuracy on  the Stroke Prediction Dataset (SPD).  We improved the model's transparency and clinical applicability by identifying three important clinical variables using LIME-based interpretability analysis: age, hypertension, and glucose levels.
Through early prediction, this study highlights how combining ensemble learning with explainable AI (XAI) can deliver highly accurate and interpretable stroke risk assessment. By enabling data-driven prevention and personalized clinical decisions, our framework has the potential to transform stroke prediction and cardiovascular risk management.

\end{abstract}


\section{Introduction}
Stroke remains a critical global health concern, ranking as the
second-leading cause of death and the third-leading cause of death.
Disability combined, with mortality rates rising particularly in developing nations. Annually, 13.7 million new stroke cases are reported,
leading to approximately 5.5 million deaths \cite{abujaber2025prediction}.
Stroke continues to be a leading cause of death and permanent disability worldwide. This cerebrovascular event happens when there is a disruption in cerebral blood flow, which results in oxygen deprivation and irreparable brain injury.  In order to mitigate negative effects and conduct appropriate treatments, early risk prediction is essential\cite{kumar2025stroke}.

Machine learning approaches have the potential to revolutionize the prediction of stroke and cardiovascular risk by revealing tiny predictive patterns in large healthcare datasets \cite{akib1}. By combining many clinical and risk criteria, these models produce individualized risk ratings that make precision treatment possible. Recent research has demonstrated the promise of machine learning in stroke prediction, with ensemble models showing strong performance.  Nevertheless, there are still few comparison studies of algorithms on actual clinical data\cite{srivastava2025enhancing}. We assess ten sophisticated machine learning algorithms for stroke prediction in this study.  Each has special advantages. For example, RF, ET, and XGB are excellent interpreters\cite{akib5}, while GB captures intricate nonlinear relationships for increased accuracy.  Adaptive hyperparameters in all models maximize predictive performance by reducing error\cite{rahman2025comparative}.
By allowing for early intervention, accurate stroke prediction lowers the incidence and severity of strokes.  Through the use of these models, we want to improve clinical judgment and patient outcomes by providing proactive, data-driven care \cite{pundkar2025transforming}.

Key contributions include:
\begin{itemize}
    \item Comparative Analysis of Data Balancing Techniques for Stroke Risk Prediction Across Multi-Source Patient Health Profiles.
    \item The best-performing algorithms (RF+ET+XGBoost) were combined to create a carefully planned hybrid ensemble model that improved prediction accuracy and robustness.
    \item Explainable AI via LIME for transparent decision-making\cite{akib2}.
\end{itemize}

\section{Literature Review}
ML and DL have been shown to be helpful in stroke prediction in recent studies, with RF emerging as a top performer.  For example, Rahman et al. \cite{rahman2023prediction} used RF to obtain 99\% accuracy, whereas Mushtaq et al. \cite{mushtaq2023machine} used SVM to report 99.5\% accuracy/F1-score, demonstrating the superiority of traditional ML over DL in this field.  Ensemble techniques such as the Dense Stacking Ensemble\cite{akib4} (DSE) by Hassan et al. \cite{hassan2024predictive} (96.6\% accuracy) and the soft voting ensemble (RF + ERT + HGB) by Srinivas et al. \cite{srinivas2023brain} (96.88\% accuracy) further highlight the resilience of hybrid approaches.  Results from DL models, like CNNs, were mixed. Bhowmick et al. \cite{bhowmick2024machine} demonstrated 100\% precision but needed validation, while Chahine et al.\cite{chahine2023machine} observed that DL had a lower AUC (0.764) than gradient-boosted trees.

Even with excellent accuracy, there are a few limitations that are consistent across research.  Aish et al. \cite{aish2024improving} and Mridha et al. \cite{mridha2023automated} have observed that class imbalance was a persistent problem that was frequently resolved using SMOTE, but at the risk of bias or overfitting.  Clinical adoption was hampered by computational costs and "black-box" interpretability\cite{akib3} issues, especially for DL models \cite{daidone2024machine}, while generalizability was restricted by small or retrospective datasets (e.g., Issaiy et al. \cite{issaiy2025machine}, Xie et al. \cite{xie2025comprehensive}).  Additionally, scalability and real-time data integration issues were noted by Bhowmick et al. \cite{bhowmick2024machine} and Kanna et al. \cite{kanna2024machine}, respectively.  Daidone and colleagues also brought up ethical issues, including privacy and bias in datasets \cite{daidone2024machine}.

The importance of external validation on a variety of datasets in bridging research-clinic gaps has been highlighted by Karim et al. \cite{karim2025optimizing} and Hasan et al. \cite{hasan2025enhancing}.  Although Mridha et al. \cite{mridha2023automated} found that explainability tools (like SHAP/LIME) increased transparency, wider use is required.  The integration of real-time systems, such as Karim et al. \cite{karim2025optimizing} MediaPipe ensemble, and multimodal data, such as imaging and clinical, as demonstrated in Chahine et al. \cite{chahine2023machine}, could improve utility.  In conclusion, resolving SMOTE's shortcomings \cite{aish2024improving} and emphasizing recall over accuracy (e.g., Abujaber et al. \cite{abujaber2025prediction}) may better meet clinical demands, as false negatives are more expensive than false positives.

\begin{table}[t]
\centering
\caption{Summary of Reviewed ML Studies on Stroke Prediction}
\label{tab:full_comparison}
\begin{tabular}{|p{1cm}|p{1.5cm}|p{2.5cm}|p{2cm}|}
\hline
\textbf{Study} & \textbf{Methods} & \textbf{Key Contributions} & \textbf{Limitations} \\
\hline

Daidone et al.\cite{daidone2024machine} & SVM, RF, CNN, DNN & CNN AUC $>$ 0.90; DNN outperforms clinical tools & Standardization issues; generalizability \\
\hline

Bhowmick et al.\cite{bhowmick2024machine} & ANN, SVM, KNN + EHR preprocessing & ANN accuracy 98.13\%; SVM AUC 0.68 & EHR bias; limited population \\
\hline

Hassan et al.\cite{hassan2024predictive} & DSE (TabNet+XGB+RF) + SMOTE & Accuracy 96.6\%; AUC 98.9\% & Complex model; high data dependency \\
\hline

Kanna et al.\cite{kanna2024machine} & RF, SVM, DT + Flask GUI & RF accuracy 94.3\%; GUI-based system & No real-time data; static features \\
\hline

Abujaher et al.\cite{abujaber2025prediction} & ANN, XGB, RF + SHAP & ANN F1-score 86\%; AUC 94\% & Single-center data; retrospective bias \\
\hline

Hasan et al.\cite{hasan2025enhancing} & XGB, RF, KNN + feature selection & XGB accuracy 99\%; AUC 100\% & Limited to binary outcomes \\
\hline

Karim et al.\cite{karim2025optimizing} & RF+XGB+CB + MediaPipe landmarks & Accuracy 94.8\%; real-time response & Performance drops in low-light \\
\hline

Issaiy et al.\cite{issaiy2025machine} & Systematic review (ML/DL) & GB AUC median 0.91; clinical support & Small samples; retrospective data \\
\hline

Xie et al.\cite{xie2025comprehensive} & RF, XGB, Cox model (CHARLS data) & C-index $>$ 0.70; key risk factors identified & Population-specific model \\
\hline

Mridha et al.\cite{mridha2023automated} & RF + SHAP, LIME + Under-sampling + SMOTE & Accuracy 90.36\%; Explainability via SHAP/LIME; Real-time web app & No external validation; high computation; no feature selection \\
\hline

Aish et al.\cite{aish2024improving} & Bagging + SMOTE & Accuracy 98.3\%; ROC AUC 99.5\%; SMOTE improved performance & No comparison with other balancing; lacks interpretability \\
\hline

Chahine et al.\cite{chahine2023machine} & Ensemble (GBT+Cox), RF, ANN + RUS/SMOTE & AUC up to 0.892; Ensemble outperforms clinical scores & Retrospective review; generalizability issues \\
\hline

Srinivas et al.\cite{srinivas2023brain} & RF, ERT, HGB + SMOTE & Accuracy 96.88\%; Soft voting ensemble & Cannot classify stroke type; no optimization \\
\hline

Mushtaq et al.\cite{mushtaq2023machine} & SVM + Random Oversampling & Accuracy 99.5\%; Specificity 99\% & Limited to tabular Kaggle data; lacks generalization \\
\hline

Rahman et al.\cite{rahman2023prediction} & RF, XGB, ANN + ROS & Accuracy 99\%; PCA used; compared ML/DL & DL underperformed; dataset lacked stroke diversity \\
\hline
\textbf{Our Ensemble} & RF, ET, XGB (ROS) on SPD dataset & Data balancing(ROS), Ensemble model, XAI; Accuracy 99.09\%; & Data quality issues, Dataset Limitations.\\
\hline
\end{tabular}
\end{table}

Table~\ref{tab:full_comparison} compares our RF+ET+GB ensemble with ROS to 15 previous studies and reveals that it performs better(SPD: 99.09\% accuracy, 99.10\% F1-Score) while addressing important clinical interpretability and data imbalance issues that impact traditional methods. Current stroke prediction models have issues with data quality, generalizability, and computing efficiency.  Our technique addresses these issues by addressing imbalances, methodically preparing data, and rigorously evaluating several ML models with an ensemble method using grid search and cross-validation.  We enhance stroke risk assessment's clinical usability and predictive accuracy by combining explainable AI with comprehensive measures (F1-score).

\section{Methodology}

\label{sec:Method}
\begin{figure*}[h]
\centerline{\includegraphics[scale=0.47]{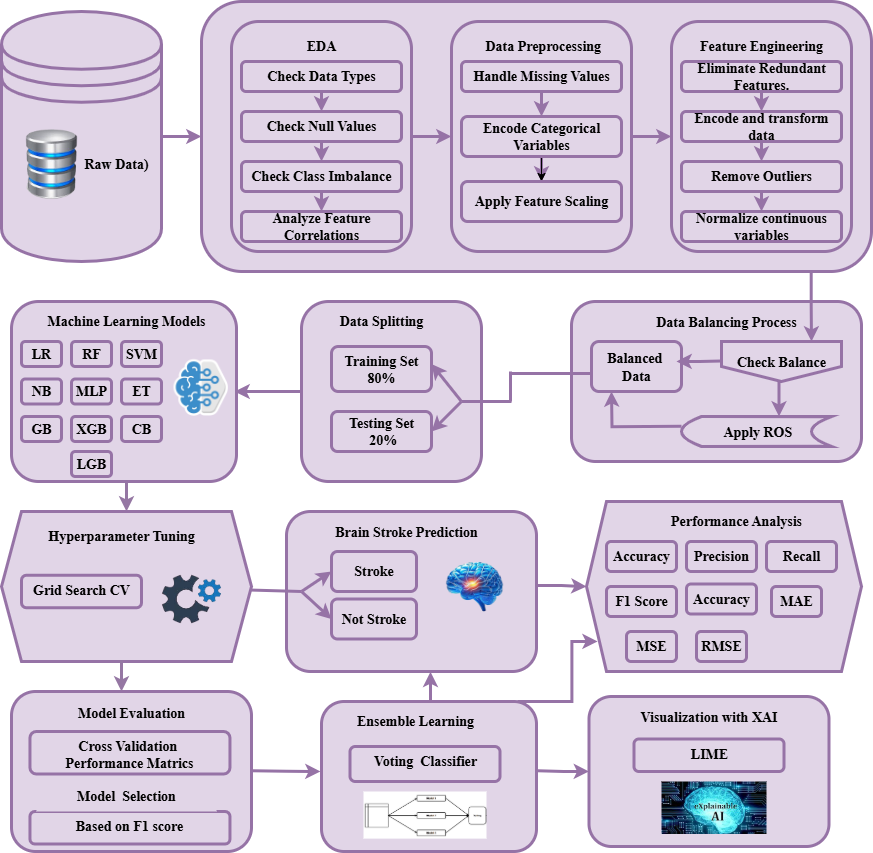}}
\caption{The Proposed Methodology Diagram for Stroke Risk Prediction.}
\label{fig:Methodology}
\end{figure*}
This study uses publicly available Stroke Prediction Datasets SPD and SDP to present improved machine learning (ML) algorithms for both high-precision stroke classification and improved stroke detection accuracy.
Our approach, which is shown in Figure \ref{fig:Methodology}, consists of multiple crucial steps meant to address class imbalance and create a reliable stroke detection model with the ensemble method.  The ensuing sections provide a detailed explanation of each stage:

\subsection{Dataset Description}
\begin{figure}[h]
    \centering
    \includegraphics[width=0.19\textwidth]{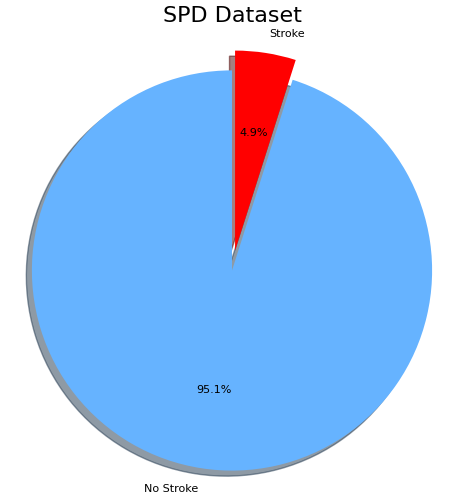}
    \hspace{0.5cm}
    \includegraphics[width=0.2\textwidth]{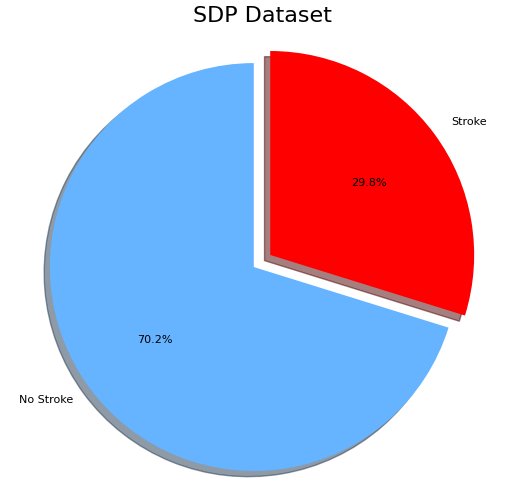}
    \hspace{0.5cm}
    \caption{The Stroke Distribution of two Datasets}
    \label{fig:StrokeDistribution}
\end{figure}

Our study integrates real-world patient data, neurological information, and synthetic symptom profiles from two anonymized databases to predict stroke risk accurately.
For a robust stroke risk prediction, this study uses two complementary datasets:

Dataset Stroke Prediction Dataset (SPD), which included lifestyle, clinical, and demographic characteristics.  Classification algorithms for early stroke risk identification in healthcare are supported by the dataset and this incorporates stroke comorbidity data into its extensive neurological profiles. The two datasets class imbalance is depicted in Figure \ref{fig:StrokeDistribution}. 

\subsection{Exploratory Data Analysis (EDA)}
Exploratory Data Analysis (EDA) was performed to assess variable distributions, detect missing values, and examine class imbalance. Correlation analysis was conducted to identify relationships among predictors and the target variable, helping guide later preprocessing and feature engineering steps.

\subsection{Data Preprocessing}
Several domain-appropriate preprocessing steps were applied to improve data quality. Missing values were handled using multivariate techniques such as KNN and Iterative Imputer for variables with complex dependencies, while mean/median imputation was used for numerical attributes and mode imputation for binary or categorical fields. Feature scaling through standardization ensured consistent magnitude across all predictors.

\subsection{Feature Engineering}
Feature engineering was implemented to enhance the predictive strength of the dataset. Dimensionality reduction techniques were applied to address multicollinearity and refine feature relevance. Additionally, outlier removal and further standardization were used to improve the stability of continuous variables. These transformations preserved essential predictive information while increasing compatibility with machine learning models.

\subsection{Data Balancing via the Random Over-Sampling}
For stroke risk prediction, Random Over-Sampling (ROS) outperformed other data balancing methods, according to our comparison research.  Conversely, ROS improved accuracy through minority class replication while maintaining data authenticity, whereas other techniques \ref{tab:different_balancing_techniques} demonstrated only moderate impact.
\begin{table}[h]
\scriptsize
\centering
\caption{Comparative Performance of Balancing Techniques on SPD and SDP Datasets (\%)}
\label{tab:different_balancing_techniques}
\begin{tabular}{|c|c|c|c|c|}
\hline
\textbf{Technique} & \multicolumn{2}{c|}{\textbf{Dataset 2}} & \multicolumn{2}{c|}{\textbf{Dataset 3}} \\
\cline{2-5}
 & Acc & F1 & Acc & F1 \\
\hline
ROS    & 98.59 & 98.62 & 84.04 & 84.72 \\ \hline
SMOTE  & 95.16 & 95.23 & 77.00 & 77.40 \\ \hline
BSMOTE & 96.03 & 96.09 & 77.00 & 77.40 \\ \hline
ADASYN & 94.93 & 94.94 & 77.00 & 77.40 \\ \hline
\end{tabular}
\end{table}

\subsection{ML algorithms}
In order to predict strokes, we used our balanced dataset to test ten machine learning models:
\begin{itemize}
    \item \textbf{Tree-based}: Gradient Boosting (GB), Random Forest (RF), ExtraTrees, XGBoost (XGB), CatBoost (CGB), and LightGBM (LGB)
    \item \textbf{Linear}: Logistic Regression (LR) 
    \item \textbf{Probabilistic}: Naïve Bayes (NB)
    \item \textbf{Non-linear}: Support Vector Machine (SVM).
    \item \textbf{Neural}: Multilayer Perceptron (MLP).
\end{itemize}
\subsection{Hyperparameter Tuning}
 Following mathematical formulations and empirical best practices, we used Grid Search CV (5-fold) to optimize all 10 ML models with algorithm-specific hyperparameter spaces.
The tuning process for each model $m \in \mathcal{M}$ (which $\mathcal{M}$ contains all 10 algorithms) follows:
\begin{equation}
\theta^*_m = \underset{\theta \in \Theta_m}{\arg\max} \left( \frac{1}{K} \sum_{k=1}^K F1\left(f_m(X^{(k)}_{train}; \theta), y^{(k)}_{train}\right) \right)
\end{equation}
where:
\begin{itemize}
    \item $\Theta_m$ is the parameter space for model $m$ (e.g., for XGBoost: $\Theta_{XGB} = \{\eta, \gamma, \lambda, \text{max\_depth}, ...\}$)
    \item $K=5$ is the number of cross-validation folds
    \item $F1 = 2 \cdot \frac{\text{precision} \cdot \text{recall}}{\text{precision} + \text{recall}}$ is our optimization metric
    \item $f_m(X; \theta)$ denotes model $m$ with parameters $\theta$ making predictions on $X$
\end{itemize}
The grid search evaluates all combinations in the Cartesian product:
\begin{equation}
\Theta_m = \prod_{i=1}^{P_m} \Theta_m^{(i)}
\end{equation}
where $P_m$ is the number of tunable parameters for the model $m$.

The table \ref{tab:all_hyperparams} defines the Key Parameters for 10 ML models used in stroke prediction.

\begin{table}[h]
\scriptsize
\centering
\caption{Key Parameters in Methodology}
\label{tab:all_hyperparams}
\begin{tabular}{l c c c}
\toprule
\textbf{Model} & \textbf{Parameter} & \textbf{Values} \\
\midrule
LR & C & 0.001, 0.01, 0.1, 1, 10 \\
   & Penalty & l1, l2 \\
   & Solver & liblinear \\ \hline

RF & n\_estimators & 100, 200 \\
   & max\_depth & None, 10, 20 \\
   & min\_split & 2, 5 \\ \hline

SVM & C & 0.1, 1, 10 \\
    & gamma & scale, auto \\
    & kernel & rbf \\ \hline

ET & n\_estimators & 100, 200 \\
   & max\_depth & None, 10, 20 \\
   & class\_wt & balanced \\ \hline

XGB & n\_est & 100, 200 \\
    & max\_depth & 3, 6, 9 \\
    & lr & 0.01, 0.1 \\ \hline

GB & n\_est & 100, 200 \\
   & lr & 0.01, 0.1 \\
   & max\_depth & 3, 6 \\ \hline

LGB & n\_est & 100, 200 \\
    & num\_leaves & 31, 63 \\
    & lr & 0.01, 0.1 \\ \hline

CB & iterations & 100, 200 \\
   & depth & 4, 6 \\
   & lr & 0.01, 0.1 \\ \hline

MLP & layers & (50), (100), (50,50) \\
    & activ & relu, tanh \\
    & alpha & 0.0001, 0.001 \\ \hline

NB & var\_smooth & 1e-9, 1e-8, 1e-7 \\
\bottomrule
\end{tabular}
\end{table}

\subsection{Model Evaluation and Selection}
Models were evaluated via stratified K-Fold CV (F1/Accuracy/AUC) and selected by F1 score to handle class imbalance effectively.

\subsection{Ensemble Model}
The probability outputs from three optimized base models (RF, ET, and XGBoost) out of ten models were combined to create a soft voting ensemble.  According to their ranks for cross-validation performance, model weights were allocated.  Weighted averaging of class probabilities is used by the ensemble to aggregate predictions, and F1 score tradeoff analysis is used to optimize decision thresholds.  By using consensus-based prediction, this method reduces the biases of individual models while utilizing the complementing strengths of various tree-based designs.

\subsection{XAI with LIME}
We examined feature significance trends in our ensemble models using Local Interpretable Model-agnostic Explanations (LIME) to improve model interpretability.  LIME works by producing local, comprehensible approximations of how the model behaves in relation to particular predictions. The most significant clinical characteristics (such as age and hypertension) were determined for each instance using thousands of perturbed samples, and their influence on risk estimates was measured.

\section{Results and Discussion }
\label{sec:Results}

\subsection{Performance Analysis}
Table \ref{tab:dataset_SPD} reveals that tree-based ensemble approaches (RF, ET, XGB) produced greater performance in stroke detection on SPD dataset, while other models showed significantly lower accuracy.

\begin{table}[h]
\scriptsize
\centering
\caption{Comparative Performance Analysis of Machine Learning Models for Stroke Risk Prediction Using Dataset SPD}
\label{tab:dataset_SPD}
\begin{tabular}{l c c c}
\toprule
\textbf{Model} & \textbf{Accuracy} & \textbf{F1 Score} & \textbf{AUC} \\
\midrule
LR   & 77.85 & 79.35 & 84.90 \\
\textbf{RF}  & \textbf{98.51} & \textbf{98.53} & \textbf{100.00} \\
SVM  & 86.93 & 87.81 & 91.62 \\
MLP  & 95.20 & 95.43 & 97.68 \\
NB   & 76.53 & 76.70 & 82.49 \\
GB   & 97.05 & 97.14 & 99.83 \\
\textbf{XGB} & \textbf{97.63} & \textbf{97.69} & \textbf{99.91} \\
CB   & 94.74 & 95.00 & 98.97 \\
LGB  & 96.88 & 96.98 & 99.78 \\
\textbf{ET}  & \textbf{99.09} & \textbf{99.10} & \textbf{100.00} \\
\bottomrule
\end{tabular}
\end{table}

In stroke detection on SDP dataset, RF and ET performed better than traditional models (LR/SVM/NB), and XGB (GB/LGB) also performed exceptionally well, demonstrating the clinical usefulness of tree-based ensembles in Table \ref{tab:dataset_SDP}.

\begin{table}[h]
\scriptsize
\centering
\caption{Comparative Performance Analysis of Machine Learning Models for Stroke Risk Prediction Using Dataset SDP}
\label{tab:dataset_SDP}
\begin{tabular}{l c c c}
\toprule
\textbf{Model} & \textbf{Accuracy} & \textbf{F1 Score} & \textbf{AUC} \\
\midrule
LR           & 76.81 & 77.50 & 84.99 \\
\textbf{RF}  & \textbf{83.92} & \textbf{84.68} & \textbf{92.23} \\
SVM          & 77.21 & 77.29 & 83.73 \\
MLP          & 77.18 & 77.64 & 85.28 \\
NB           & 75.46 & 76.69 & 84.42 \\
GB           &  81.91 & 82.70 & 89.75 \\
\textbf{XGB} & \textbf{82.49} & \textbf{83.26} & \textbf{89.84} \\
CB           & 78.72 & 79.04 & 87.33 \\
LGB          & 81.85 & 82.66 & 89.62 \\
\textbf{ET}           & \textbf{84.03} & \textbf{84.57} & \textbf{92.96} \\
\bottomrule
\end{tabular}
\end{table}

\subsection{Ensemble Model Comparison}

The ensemble model demonstrated exceptional performance on the SPD dataset, achieving 99.09\% accuracy and perfect recall and AUC values, as shown in Table~\ref{tab:ensemble_cls}. 
The SDP dataset also provided strong and stable performance (84.04\% accuracy, 92.57\% AUC), although with comparatively higher error values (MAE = 0.18), as presented in Table~\ref{tab:ensemble_err}. 
These results highlight the robustness of the ensemble model across diverse clinical datasets.

\begin{table}[h]
\scriptsize
\centering
\caption{Classification Metrics of the Ensemble Model}
\label{tab:ensemble_cls}
\setlength{\tabcolsep}{5pt}
\renewcommand{\arraystretch}{1.15}
\begin{tabular}{lccccc}
\toprule
\textbf{Dataset} & \textbf{Acc (\%)} & \textbf{Prec (\%)} & \textbf{Rec (\%)} &
\textbf{F1 (\%)} & \textbf{AUC (\%)} \\
\midrule
SPD & 99.09 & 98.22 & 100.00 & 99.10 & 100.00 \\
SDP & 84.04 & 82.14 & 87.47 & 84.72 & 92.57 \\
\bottomrule
\end{tabular}
\end{table}

\begin{table}[h]
\scriptsize
\centering
\caption{Error Metrics of the Ensemble Model}
\label{tab:ensemble_err}
\setlength{\tabcolsep}{6pt}
\renewcommand{\arraystretch}{1.15}
\begin{tabular}{lccc}
\toprule
\textbf{Dataset} & \textbf{MAE} & \textbf{MSE} & \textbf{RMSE} \\
\midrule
SPD & 0.04 & 0.01 & 0.11 \\
SDP & 0.18 & 0.05 & 0.22 \\
\bottomrule
\end{tabular}
\end{table}

\subsection{Ablation Study}
Prediction quality was significantly enhanced by data balancing, as seen in Table \ref{tab:ablation_study} by SPD reaching nearly perfect scores (99.69\% R2). All datasets demonstrated an error reduction of 85-98\%, despite SDP demonstrating modest results (+ 85.11\% R²), confirming the crucial impact of preprocessing in clinical machine learning.
\begin{table}[h] 
\scriptsize
\centering
\caption{Impact of Data Balancing \& Tuning on Stroke Prediction}
\label{tab:ablation_study}

\begin{tabular}{@{}lccccc@{}}
\toprule
\textbf{Dataset} & \textbf{Setup} & \textbf{MSE} & \textbf{MAE} & \textbf{R²} \\
 & & \scriptsize{(Train/Test)} & \scriptsize{(Train/Test)} & \scriptsize{(Test)} \\ \midrule

\multirow{2}{*}{SPD}
& Unbalanced & 0.04/0.04 & 0.09/0.09 & 0.07 \\
& Balanced & 0.00/0.00 & 0.00/0.06 & 99.69\% \\ \addlinespace

\multirow{2}{*}{SDP}
& Unbalanced & 0.15/0.14 & 0.32/0.31 & 0.31 \\
& Balanced & 0.03/0.05 & 0.10/0.18 & 85.42\% \\ \bottomrule
\end{tabular}
\end{table}

\subsection{XAI Analysis}
\begin{figure}[h]
    \centering
    \includegraphics[scale=0.42]{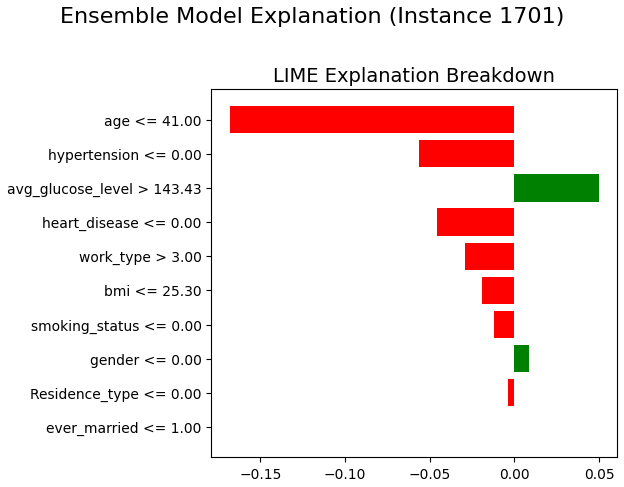}
    \caption{LIME Explanation for Ensemble Model Using Dataset SPD.}
    \label{fig:LIME_SPD}
\end{figure}
\begin{figure}[h]
    \centering
    \includegraphics[scale=0.42]{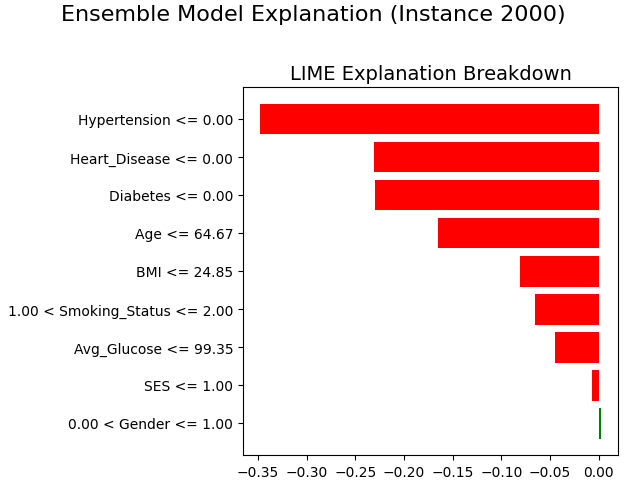}
    \caption{LIME Explanation for Ensemble Model Using Dataset SDP.}
    \label{fig:LIME_SDP}
\end{figure}
Our XAI analysis (Figs. \ref{fig:LIME_SPD} and \ref{fig:LIME_SDP}) uses LIME to illustrate the feature‐importance patterns across the SPD and SDP datasets. In both cases, major clinical factors—such as age and hypertension—show dominant influence on model predictions. This interpretability validation reinforces clinical confidence in the ensemble model.

\section{Conclusion}
In summary, Our research used an RF+ET+XGB ensemble with XAI to create an interpretable stroke prediction system that achieved 84.04\% (SDP) and 99.09\% (SPD) accuracy.  Age, hypertension, and glucose were validated as important clinical predictors by LIME/SHAP analysis, which showed excellent performance and interpretability for clinical use.

The study faced significant challenges and limitations in spite of these developments.  The quality of the data, especially the absence of clinical information, was a major obstacle to efficient feature selection.  Although useful, the use of publicly accessible datasets had limitations in terms of size and diversity, indicating that larger, more diverse real-world cohorts are needed to validate broader generalizability.  Additionally, class imbalance was successfully handled by Random Over-Sampling; nevertheless, the underlying complications of this problem, such as the potential for bias or overfitting with methods like SMOTE, still apply.  There are also continuous considerations regarding practical barriers to clinical use, such as computing expenses and the "black-box" character of many sophisticated models.

Future research will concentrate on a few crucial areas to expand on these discoveries.  To improve generalizability and practical applicability, the following areas should be prioritized for future directions: (1) multicenter data integration; (2) deep learning-enhanced feature extraction; and (3) cloud-based clinical deployment platforms.  Through a robust, interpretable stroke prediction system, this study bridges the gap between AI and clinical practice, enabling data-driven preventative techniques to lower healthcare costs through early, personalized risk assessment.

\bibliographystyle{IEEEtran}
\bibliography{sample}

\end{document}